# An Alternate Approach for Designing a Domain Specific Image Search Prototype Using Histogram

Sukanta Sinha[1], Rana Dattagupta[2], Debajyoti Mukhopadhyay[1,3]

[1]Web Intelligence and Distributed Computing Research Lab, Kolkata, India; [2]Computer Science Department, Jadavpur University, Kolkata, India; [3]Information Technology Department, Maharashtra Institute of Technology, Pune, India.
Email: sukantasinha2003@gmail.com, ranadattagupta@yahoo.com, debajyoti.mukhopadhyay@gmail.com

## ABSTRACT

Everyone knows that thousand of words are represented by a single image. As a result image search has become a very popular mechanism for the Web-searchers. Image search means, the search results are produced by the search engine should be a set of images along with their Web-page Unified Resource Locator (URL). Now Web-searcher can perform two types of image search, they are "Text to Image" and "Image to Image" search. In "Text to Image" search, search query should be a text. Based on the input text data system will generate a set of images along with their Web-page URL as an output. On the other hand, in "Image to Image" search, search query should be an image and based on this image system will generate a set of images along with their Web-page URL as an output. According to the current scenarios, "Text to Image" search mechanism always not returns perfect result. It matches the text data and then displays the corresponding images as an output, which is not always perfect. To resolve this problem, Web researchers have introduced the "Image to Image" search mechanism. In this paper, we have also proposed an alternate approach of "Image to Image" search mechanism using Histogram.

## 1. Introduction

Nowadays people are uploading the number of images in the internet [1-4]. As a result internet has become a huge reservoir of digital images. Exponentially increasing digital image database volume has urged many researchers for developing effective image retrieval methods. Considering the importance of the problem, various researchers have been carried out on image search over the past few years and all are available in the literature. There are two types of image search available such as "Text to Image" search and "Image to Image" search. The "Text to Image" search mechanism expects a search text as a search query and following that search text, the search prototype will generate a set of images. "Text to Image" search approach, mainly matches the image tag information like image name, meta-tag information, etc. Now, consider such a situation where users upload their images with irrelevant text information. That time this approach will not work properly. In "Image to Image" search mechanism, search query itself contains the search image and based on the search image, search prototype will generate a set of images as an output. Unlike retrieval of "Text to Image", image search is difficult and has involved image analysis. In this paper, an attempt has been made to design a methodology for a domain specific image search prototype using histogram. This prototype fully deals with "Image to Image" search and it is a domain specific approach.

The paper is organized in the following way. In Section 2, the basics of histogram and gray-scale of an image are discussed. Section 3 discusses the existing works. The proposed architecture for domain specific image search prototype using histogram is discussed in section 4. All the components of our architecture have been also discussed in the same section. The experimental analysis and the conclusions reached from our study have been summarized in sections 5 and 6.

## 2. Histogram Basics

Image histograms are an important concept in Image Processing [5-9]. The histogram of an image refers to the histogram of the intensity values of the pixels. Histogram displays the number of pixels in an image for a particular intensity level. A histogram is a graphical representation of data distribution. An image histogram is a type of histogram that represents a digital image tonal data distribution. It plots the number of pixels for each tonal value. By looking at the histogram for a specific image a viewer will be able to judge the entire tonal distribution at a glance.

### 2.1. Why Choose Histogram?

Web-pages have contained various types of images. It may be color image, gray-scale image, black-and-white image or various types of image formats like .jpg, .jpeg, etc. To perform an image search, we need to make all the images in a common format. The histogram is one of the simple and useful tool to process those images and produce a common format.

### 2.2. What do You Mean by Gray-Scale Image?

A gray-scale digital image is an image in which the value of each pixel is a single sample, that is, it carries only intensity information. Images of this sort, also known as black-and-white, are composed exclusively of shades of gray, varying from black at the weakest intensity to white at the strongest [10].

### 2.3. Histogram of a Gray-Scale Image

The gray-scale histogram of an image represents the distribution of the pixels in the image over the gray-level scale. It can be visualized as if each pixel is placed in a bin corresponding to the color intensity of that pixel. All of the pixels in each bin are then added up and displayed on a graph. This graph is the histogram of the image.

## 3. Existing Work

Image search is such a complex mechanism, where various researches are going on to improve the search prototype. Our paper is not intended to provide a complete survey of techniques. According to our knowledge, we have applied these techniques on few examples. Now a day's research on search engine has been carried out in universities and open laboratories, many dot-com companies. Unfortunately, many of these techniques are used by dot-coms, and especially the resulting performance, are kept private behind company walls, or are disclosed in patents that can be comprehended and appreciate by the lawyers. Therefore, we believe that the overview of problems and techniques that we presented here can be useful.

In this section, we have explained few existing mechanisms and explained how the current systems are working.

### 3.1. Definitions

- Ontology –It is a set of domain related key information, which is kept in an organized way based on their importance.
- Relevance Value –It is a numeric value for each Web-page, which is generated on the basis of the term Weight value, term Synonyms, number of occurrences of Ontology terms which are existing in that Web-page.
- Seed URL –It is a set of base URL from where the crawler starts to crawl down the Web pages from the Internet.
- Weight Table – This table has two columns, first column denotes Ontology terms and second column denotes weight value of that Ontology term. Ontology term weight value lies between '0' and '1'.
- Syntable - This table has two columns, first column denotes Ontology terms and second column denotes synonym of that ontology term. For a particular ontology term, if more than one synonym exists, those are kept using comma (,) separator.
- Relevance Limit –It is a predefined static relevance cut-off value to recognize whether a Web-page is domain specific or not.
- Term Relevance Value – It is a numeric value for each Ontology term, which is generated on the basis of the term Weight value, term Synonyms, number of occurrences of that Ontology term in the considered Web-page.

### 3.2. Domain Specific Crawling

Domain specific crawling means the Web crawler crawls only domain specific Web-pages [11-18]. For finding domains based on the Web-page content, first parsed the Web-page content and then extracted all the Ontology terms as well as syntable terms [19-22]. Then each distinct Ontology term was multiplied with their respective Ontology term weight value. Ontology term weight values are taken from weight table. In this approach, for any syntable term used corresponding Ontology term weight value. Finally, taken a summation of these individual terms weight value and this value is called relevance value of that Web-page.

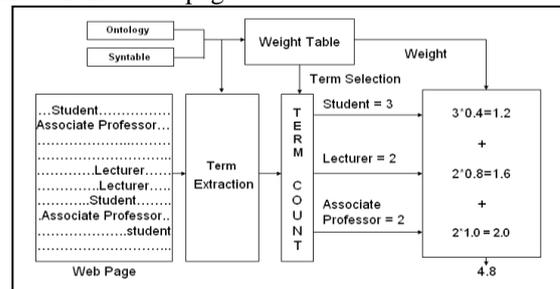

**Figure 1. Web-page Relevance Calculation Mechanism.**

Now if this relevance value is greater than the predefined "Relevance Limit" of that domain, then that Web-page belongs to a predefined particular domain otherwise discard the Web-page, i.e., the Web-page didn't belong to our domain. In **Figure 1** we have shown a mechanism to find a domain based on the Web-page content. Here, we consider 'computer science' Ontology, syntable and weight table of computer science Ontology for finding a Web-page belongs to the computer science domain or not. Suppose, the considered Web-page contains 'stu-dent' term 3 times, 'lecturer' term 2 times and 'associate professor' term 2 times and student, lecturer and associate professor weight values in the computer science domain are 0.4, 0.8 and 1.0 respectively. Then the relevance value becomes (3*0.4 + 2*0.8 + 2*1.0) = 4.8. Now, if 4.8 is greater than the relevance limit, then we called the considered Web-page belongs to the computer science domain otherwise we discard the Web-page.

### 3.3. Existing Text to Image Search

In "Text to Image" search mechanism Web searcher will provide a search text and based on that text system will find the images [23-26]. Let me explain by taking an example where this mechanism will not work. Consider a Web user (A) was created a "facebook" profile using Sachin Tendulkar's image, which is an invalid image with respect to the user (A). Now another Web user (B) wants to see the 'A' user image that time if 'B' user performs a text to image search based on A's user name. The search result will produce Sachin Tendulkar's image not 'A' user image. Hence it is an invalid search result for 'B' user's side. To resolve this problem we need to analyze the images and then only we can produce correct search results. For that reason Web researcher are introduced "Image to Image" search mechanism which explained in the next subsection.

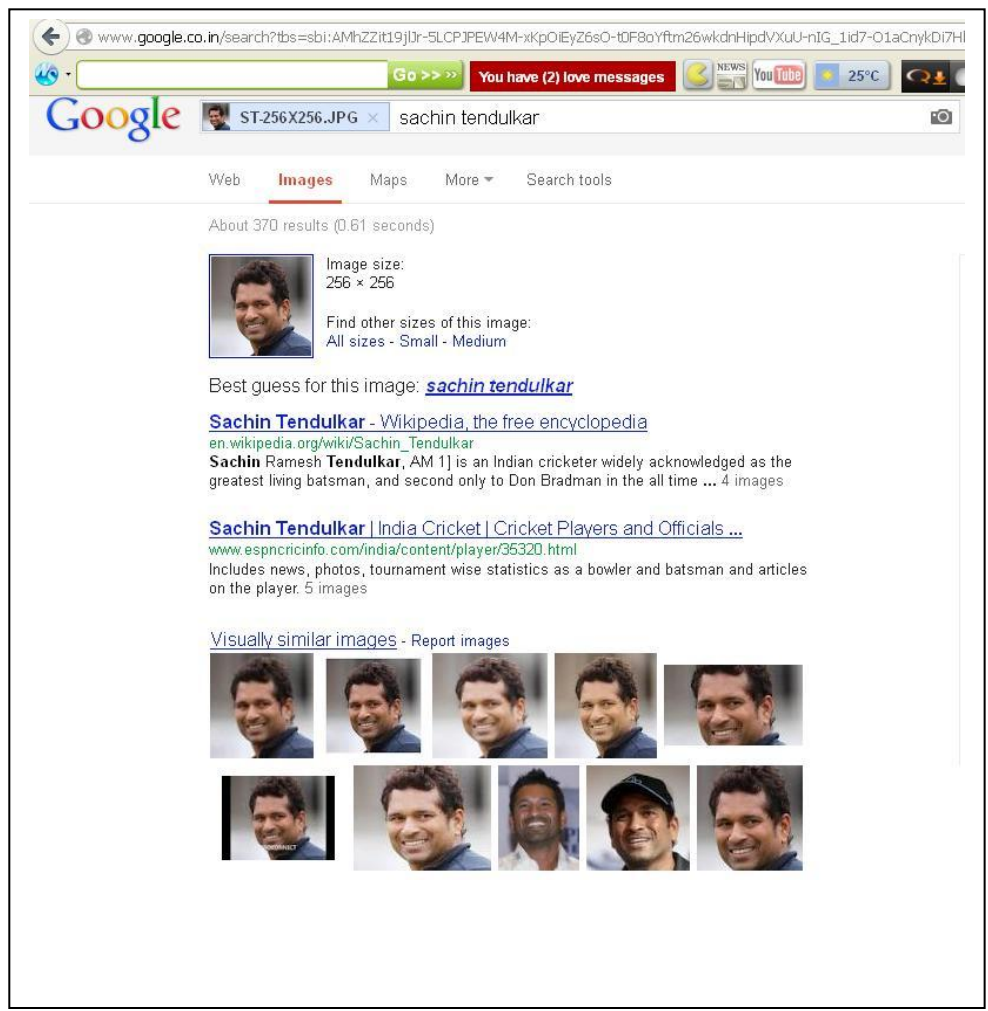

**Figure 2. Image search for a popular image (Sachin Tendulkar).**

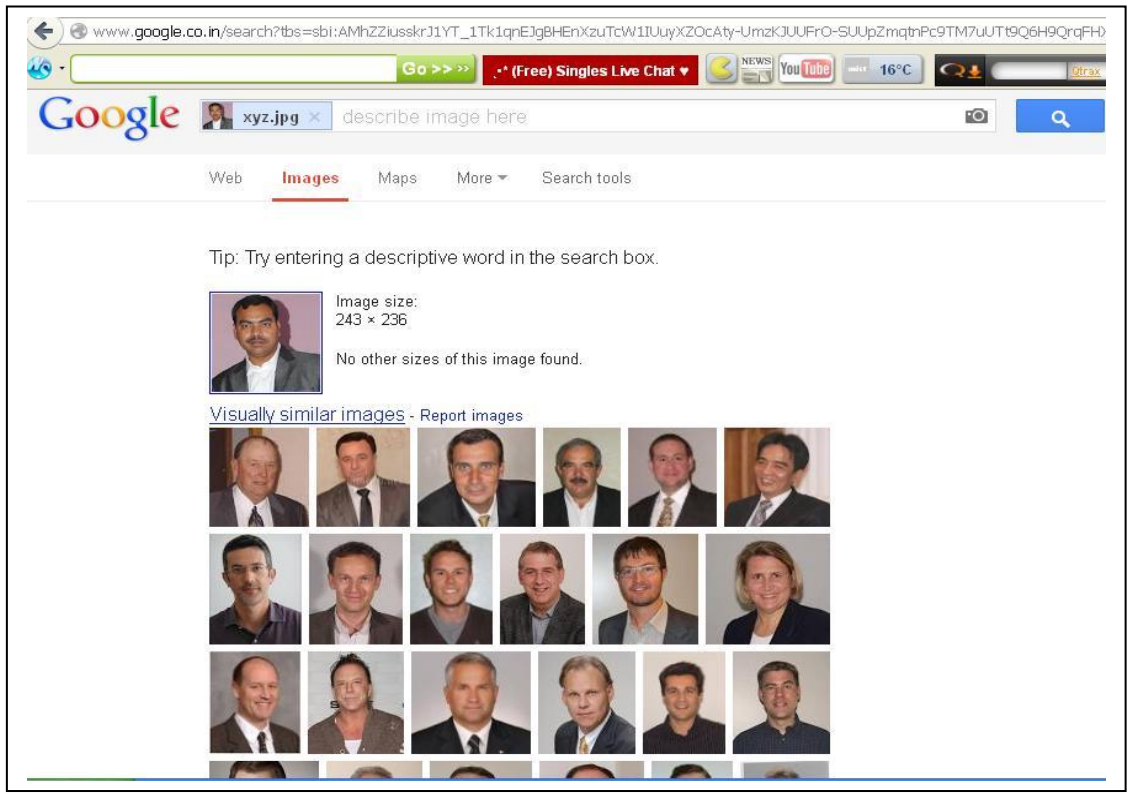

Figure 3. Image search for a non-popular image (xyz).

### 3.4. Existing Image to Image Search

In "Image to Image" search Web searcher gives an image as a search query [27-29]. Now there are various types of images such as color image, gray-scale image, black-and-white image and different image file formats like .jpg, .jpeg, .bmp, .tif, etc are available in the internet. We have done a survey where we found some issues. For popular images like 'Rabindranath Tagore', 'Sachin Tendulkar', etc. , the search engines are working fine (refer **Figure 2**) but those images which are not a popular image that time we have received a lot of irrelevant results. Suppose we have a 'xyz' image, which is not a popular image but available in few websites like 'facebook', 'LinkedIn', etc. Using this 'xyz' image while we performing the search operation, we have not received relevant results (refer **Figure 3**). We have also found another issue, say I have a color image and lots of Web-pages are exists in the internet which holding the same image but they are black-and-white image. In this scenario, if we put a color image as a search image and there does not exist any Web-page which contains the same color image that time also we found a lot of irrelevant results. As a result the Web users were misguided. To resolve this problem we have proposed an alternate image search method using histogram which can be useful and described in successive sections.

## 4. Our Approach

There are various approaches followed by the Web researchers for searching an image from the internet. We also proposed a unique alternate method for image searching and we believe that this approach can be useful. Dealing with images is always a huge challenge. There are two major difficulties we have faced. In the internet lot of images are exists and there we have found some similar images with different size. Those images are present in different Web-pages, i.e., Web-page URLs are different. In that case we have used % of pixel distribution to resolve the various types of image size used. The "Percentage (%) of pixels distribution" calculation logic has given in Equation (1).

Percentage (%) of Pixels exist in $x_i$ th position ($P(x_i)$) =
$(\lambda / TNP) *100 \%$ (1)

Where, $0 \leq i \leq 255$; TNP = Total number of pixels exist in the image; $\lambda$= Number of pixels found in $x_i$ th position of the histogram

Internet is a conglomeration of huge amount of multi-colored images. To avoid the color variations, we have converted all the images into 8-bit gray-scale image. In our approach, we have divided our proposed process into small modules like "generation of image repository", "Search result generation", etc. In subsequent sections, we have explained these modules.

### 4.1. Image Repository Creation

Image repository creation is an important role to perform "Image to Image" search. Our proposed algorithm has given below:

Input    : Seed URLs, Weight Table, Syntable
Output   : Image repository

Step 1: Initialize the crawlers by the Seed URLs.
Step 2: Crawl the Web-pages from internet.
Step 3: Calculate relevance value.
Step 4: if Relevance Value > Relevance Limit then
   If Web-page contains any image then
      i) Get the image and save the image.
      ii) Convert the image into 8-bit gray scale image.
      iii) Generate histogram of that gray scale image.
      iv) Calculate percentage (%) of pixels distribution using Equation (1).
      v) Save all the % of pixels distribution.
Step 5: go to step-2 until crawler has no URL.
Step 6: End

In **Figure 4**, we have shown the image repository creation mechanism. Before running the crawler, we need to set the seed URLs, weight table and syntable. We have used MATLAB 7.10 for generating histograms. First crawler crawls the domain specific Web-pages from the internet. Then identify the Web-pages which contain images and get the images for further processing. Connect the images into 8-bit gray-scale image and generate histograms. Now using Equation (1) calculates percentage of pixel distributions and save it for producing search results.

### 4.2. Search Result Generation Mechanism

To produce search results, we need to analyze the search image and then matched with our repository image attributes. We are basically matching all $P(x_i)$ where 'i' belongs to [0, 255] and the matching dine between search image and our repository image (refer **Figure 5**). Now the basic concept of our proposed search result generation mechanism is divided into two parts one exact match images and probable matched images. In exact match search mechanism, users will receive those Web-pages as a search result where search image was exactly matched with repository images except their size and color, because we are proposing such a prototype where image size and color has no impact on producing search results. In probable match search mechanism, we have used a tolerance value to match the percentage (%) of distributions and then produced the search results. Tolerance value is a user given number. For exact match case it is zero (0) and other cases it must be a positive integer number in the range of [0, 100].

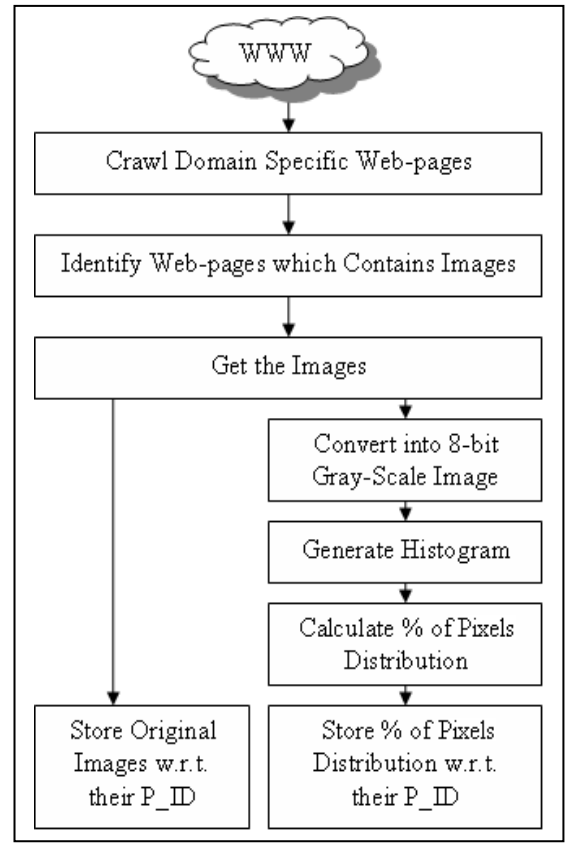

**Figure 4. Image repository generation.**

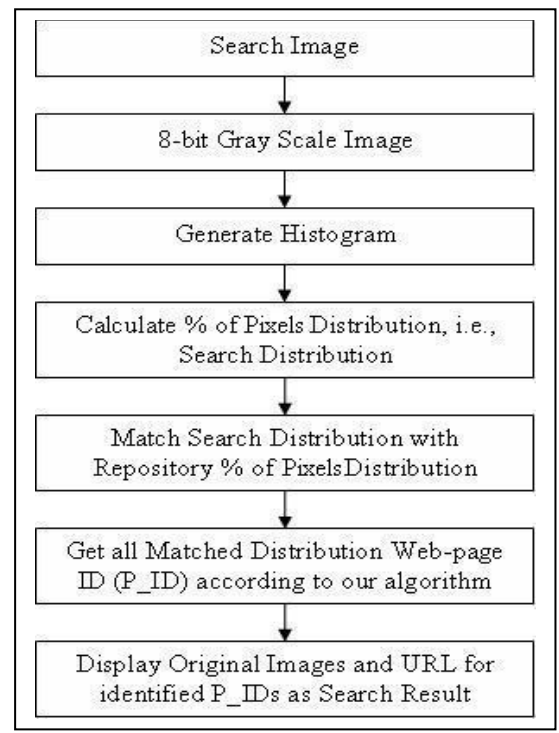

**Figure 5. Proposed search mechanism.**

### 4.3. User Interface

**Figure 6** shows a part of the user interface of our image search prototype. An image URL is typed or browses in the input image URL box, select either exact image or probable image match radio button, enter tolerance value, select the domain and enter the relevance range. In the user interface mandatory fields are denoted by star (*) sign. Relevance range default value set as [maximum relevance value, minimum relevance value], which is an editable field and according to the requirement, user can customize the relevance range values. We are providing flexibility to the users to get exact matched images or probable matched images by selecting the radio buttons in the user interface. While we select exact match that time tolerance value field become read only and set as zero (0) on the other hand if user selects probable image match radio button in the user interface that time tolerance value field becomes editable and defaulted as zero (0). In the user interface, the maximum relevance value and minimum relevance value are set dynamically according to the user selected domain. While refreshing the database, maximum relevance value is taken using ceiling function for the largest Web-page's relevance value and minimum relevance value is taken using floor function for the smallest Web-page's relevance value. Here, three domains such as "Cricket", "Hockey", and "Foot Ball" have been considered.

**Figure 6. A Part of User Interface.**

For domain selection, we have used radio button because at a time only one domain can be selected. User first input all the necessary inputs and then click on 'Search' button of the user interface, that time our proposed search result generation mechanism called to produce the search results.

## 5. Experimental Analysis

In this section we have shown our test settings, how the images were analyzed and some sample image search results.

### 5.1. Test Settings

In this subsection we will describe different parameter settings to crawl domain specific Web-pages and their images. To run the crawler, we need to set the seed URLs, weight table and syntable. We have shown few sample data in **Table 1, Table 2** and **Table 3**.

**Table 1. Seed URLs.**

| Seed URLs |
| --- |
| http://icc-cricket.yahoo.com/ |
| http://www.cricketnext.com/index.html |
| http://www.in.com |
| http://www.cricketworld.com/ |

**Table 2. Weight table.**

| Ontology terms | Weight value |
| --- | --- |
| cricket | 0.9 |
| wicket keeper | 0.8 |
| umpire | 0.4 |
| bat | 0.2 |
| match | 0.1 |

**Table 3. Syntable.**

| Ontology terms | Synterms |
| --- | --- |
| match | competition,contest |
| stamp | stick,wicket |
| ball | conglobate,conglomerate |
| umpire | judge,moderator,referee |
| catch | capture |

### 5.2. Image Analysis

In this subsection, we have shown the image analysis and given a partial percentage of the pixels distribution chart. Our main aim is, for a single image, color and size do not affect on image searching. We have shown six images in **Figure 7. (a), (c), (e), (g), (i), (k)**. We have considered those six figures such a way where, two same color images with different size, two same black & white images with different size and two gray scale images with different size. Now from those images we have generated histogram (refer **Figure 7. (b), (d), (f), (h), (j), (l)**) and then using Equation (1), we have generated **Table 4**. From the table, we have seen color and size doesn't matter because $P(x_i)$ are holding approximately same value.

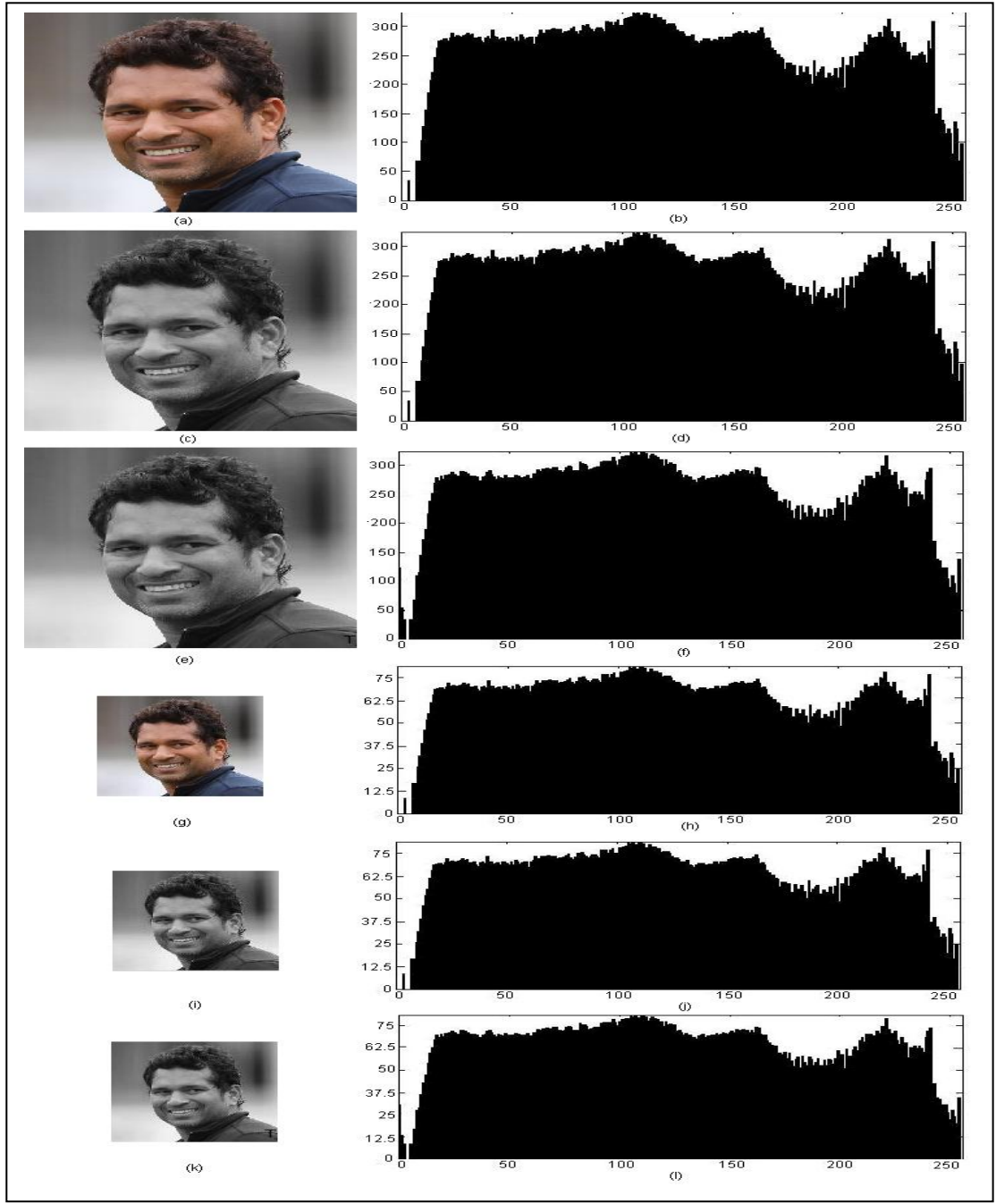

**Figure 7.(a) 256X256 Color image (b) 256X256 color image histogram after gray-scale conversion (c) 256X256 gray-scale image (d) 256X256 gray-scale image histogram (e) 256X256 black-and-white image (f) 256X256 black-and-white image histogram (g) 128X128 Color image (h) 128X128 color image histogram after gray-scale conversion (i) 128X128 gray-scale image (j) 128X128 gray-scale image histogram (k) 128X128 black-and-white image (l) 128X128 black-and-white image histogram**

Table 4. Percentage(%) of pixels distribution chart.

| Histogram Reference Figure | Percentage (%) of pixels exist in various $x_i$ th position | | | | | | | | | |
|---|---|---|---|---|---|---|---|---|---|---|
| | $P(x_{25})$ | $P(x_{50})$ | $P(x_{75})$ | $P(x_{100})$ | $P(x_{125})$ | $P(x_{150})$ | $P(x_{175})$ | $P(x_{200})$ | $P(x_{225})$ | $P(x_{250})$ |
| Figure 7(b) | 0.435 | 0.427 | 0.443 | 0.473 | 0.443 | 0.428 | 0.366 | 0.344 | 0.435 | 0.205 |
| Figure 7(d) | 0.435 | 0.427 | 0.443 | 0.473 | 0.443 | 0.428 | 0.366 | 0.344 | 0.435 | 0.205 |
| Figure 7(f) | 0.434 | 0.428 | 0.443 | 0.473 | 0.444 | 0.427 | 0.365 | 0.343 | 0.436 | 0.206 |
| Figure 7(h) | 0.436 | 0.427 | 0.444 | 0.472 | 0.443 | 0.427 | 0.366 | 0.343 | 0.436 | 0.206 |
| Figure 7(j) | 0.436 | 0.427 | 0.444 | 0.472 | 0.443 | 0.427 | 0.366 | 0.343 | 0.436 | 0.206 |
| Figure 7(l) | 0.433 | 0.427 | 0.443 | 0.473 | 0.442 | 0.426 | 0.366 | 0.345 | 0.435 | 0.205 |

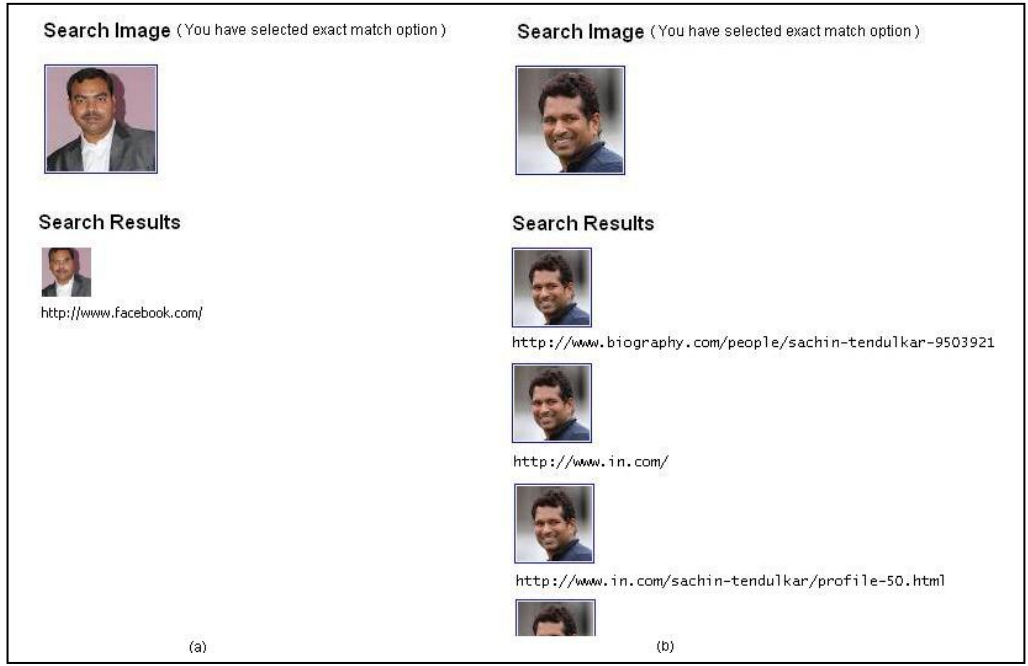

Figure 8. Image search result (a) Non-popular image (b) Popular image.

### 5.3. Image Search Results

We have used a domain specific approach, so that lots of unwanted images are already eliminated from the repository in crawling phase. As discussed in section 3.4, we have faced some issues while performing an image search in existing search prototype. Same issues we can resolve using our prototype. We have considered same search image and produced the search results. In **Figure 8**, we have shown our prototype search results. From user interface we have selected the exact match option while performed the search operation.

### 6. Conclusion

In this paper, we have proposed an alternate approach for designing a domain specific image search prototype using the histogram. Our prototype is mainly designed for few domains. Moreover, this prototype is highly scalable. We can expand supporting domains by introducing new domain Ontology and other details such as weight table, syntable, etc. Our prototype image search doesn't depend on search image color and size. We have created a uniform color and size image repository using histogram and Equation (1) percentage of pixel distribution mechanism. We have facilitated exact match and probable match option to the Web searchers using tolerance value. We are giving a wide area to the Web searchers by taking tolerance value from the user interface. While doing the survey of existing works, we found there are lot of "Image to Image" searches are going on behind the dot-com companies. But still we have found some issues on image to image search, which we can resolve using our prototype.

# REFERENCES

[1] T. Berners-Lee and M. Fischetti, "Weaving the Web: The Original Design and Ultimate Destiny of the World Wide Web by its Inventor," HarperBusiness, New York, 1999.

[2] W. Willinger, R. Govindan, S. Jamin, V. Paxson, and S. Shenker, "Scaling phenomena in the Internet," *Proceedings of the National Academy of Sciences*, New York, 19 February, 2002, pp. 2573–2580.

[3] J. J. Rehmeyer, "Mapping a Medusa: The Internet Spreads Its Tentacles," *Science News*, Vol. 171, No. 25, 2007, pp. 387-388. doi:10.1002/scin.2007.5591712503

[4] B. M. Leiner, V. G. Cerf, D. D. Clark, R. E. Kahn, L. Kleinrock, D. C. Lynch, J. Postel, L. G. Roberts and S. Wolff, "A Brief History of Internet," *ACM Computer Communication*, Vol. 35, No. 1, 2009, pp. 22-31. doi:10.1145/1629607.1629613

[5] X. Wang, J. Wu and H. Yang, "Robust Image Retrieval Based on Color Histogram of Local Feature Regions," *Springer*, Berlin, 2009.

[6] C. L. Novak and S. A. Shafer, "Anatomy of a Color Histogram," *Proceedings of IEEE Computer Society Conference on Computer Vision and Pattern Recognition*, Champaign, 15-18 June 1992, pp. 599-605. doi:10.1109/CVPR.1992.223129

[7] J. R. Smith, "Integrated Spatial and Feature Image Systems: Retrieval, Analysis and Compression," Ph.D. Thesis, Columbia University, New York, 1997.

[8] E. A. Bashkov and N. S. Kostyukova, "Effectiveness Estimation of Image Retrieval by 2D Color Histogram," *Journal of Automation and Information Sciences*, Vol. 8, No. 11, 2006, pp. 74-80.

[9] E. A. Bashkov and N. S. Shozda, "Content-Based Image Retrieval Using Color Histogram Correlation," *Graphicon proceedings*, Nizhny Novgorod, 16-21 September, 2002, pp 458-461.

[10] L. Vincent, "Morphological Grayscale Reconstruction in Image Analysis: Applications and Efficient Algorithms," *IEEE Transactions on Image Processing*, Vol. 2, No. 2, 1993, pp. 176-201.

[11] D. Mukhopadhyay, A. Biswas and S. Sinha, "A New Approach to Design Domain Specific Ontology Based Web Crawler," *The 10th International Conference on In-formation Technology*, Orissa, 17-20 December 2007, pp. 289-291.

[12] S. Chakrabarti, M. Berg, B. E. Dom, "Focused Crawling: a New Approach to Topic-specific Web Resource Discovery," *Proceedings of the 8th International World Wide Web Conference*, Elsevier, Toronto, 11-14 May, 1999, pp. 545-562

[13] D. Bergmark, C. Lagoze and A. Sbityakov, "Focused Crawls, Tunneling, and Digital Libraries," *Proceedings of the European Conference on Digital Libraries*, Rome, 16-18 September, Vol. 2458, 2002, pp. 91-106.

[14] M. Diligenti, F. Coetzee, S. Lawrence, C. L. Giles, M. Gori, "Focused crawling using context graphs," *The 26th International Conference on Very Large Databases*, Cairo, 10-14 September, 2000, pp. 527–534.

[15] N. Tyagi and D. Gupta, "A Novel Architecture for Domain Specific Parallel Crawler," *Indian Journal of Computer Science and Engineering*, Vol. 1 No. 1, 2008, pp.44-53.

[16] A. Kundu, R. D. Dattagupta and D. Mukhopadhyay, "Mining the Web with Hierarchical Crawlers-A Resource Sharing Based Crawling Approach," *International Journal on Intelligent Information and Database Systems*, Vol. 3, No. 1, 2009, pp. 90-106. doi:10.1504/IJIIDS.2009.023040

[17] A. Kundu, R. Dutta, D. Mukhopadhyay, Y. C. Kim, "A Hierarchical Web Page Crawler for Crawling the Internet Faster," *In Proceedings of the International Conference on Electronics and Information Technology Convergence*, Republic of Korea, 20 December, 2006, pp.61-67.

[18] D. Mukhopadhyay, S. Mukherjee, S. Ghosh, S. Kar, Y. C. Kim, "Architecture of A Scalable Dynamic Parallel Web Crawler with High Speed Downloadable Capability for a Web Search Engine," *In the Proceedings of the 6th International Workshop*, Republic of Korea, 20 November, 2006, pp.103-108.

[19] A. Gangemi, R. Navigli, P. Velardi, "The OntoWordNet Project: Extension and Axiomatization of Conceptual Relations in WordNet," *Proceedings of International Conference on Ontologies, Databases and Applications of Semantics*, Catania, 3-7 November, 2003, pp. 820-838.

[20] D. N. Antonio, M. Michele, N. Roberto, "A Software Engineering Approach to Ontology Building," *Information Systems*, Vol 34 No 2, 2009, pp. 258–275. doi:10.1016/j.is.2008.07.002

[21] T. Gruber, "Toward Principles for the Design of Ontologies Used for Knowledge Sharing," *International Journal of Human-Computer Studies*, Vol. 43, No. 5-6, 1995, pp. 907–928. doi:10.1006/ijhc.1995.1081

[22] M. Ehrig, and A. Maedche, "Ontology-focused crawling of web documents", *Proceedings of the 2003 ACM Symposium on Applied Computing*, Melbourne, 9-12 March, 2003, pp. 1174-1178

[23] B. Luo, X. G. Wang and X. O. Tang, "World Wide Web Based Image Search Engine Using Text and Image Con-tent Features," *Proceedings of SPIE Electronic Imaging*, Vol. 5018, Santa Clara, 20 January 2003, pp. 123-130.

[24] Z. Su, H. Zhang, S. Li and S. Ma, "Relevance Feedback Content-Based Image Retrieval: Bayesian Framework, Feature," *IEEE Transactions on Image Processing*, Vol. 12, No. 8, 2003, pp. 924-937.

[25] K.P. Yee, K. Swearingen, K. Li, M. Hearst, "Faceted Metadata for Image Search and Browsing," *CHI 2003 Proceedings*, Fort Lauderdale, 5-10 April, 2003, pp. 401-408.

[26] C. W. Niblack, R. Barber, W. Equitz, M. Flickner, E. Glasman, D. Pektovic, P. Yanker, C. Faloutsos and G. Taubin, "The QBIC Project: Querying Images by Content Using Color, Texture, and Shape," *Proceedings of Storage and Retrieval for Image and Video Databases*, San Jose, 31 January 1993, pp. 173-187.

[27] D. N. D. Harini and D. L. Bhaskari, "Image Retrieval System Based on Feature Extraction and Relevance Feedback," *Proceedings of the CUBE International In-formation Technology Conference*, Pune, 3-5 September 2012, pp.69-73.

[28] A. K. Jain and A. Vailaya, "Image Retrieval using Color and Shape," *Pattern Recognition*, Vol. 29, No. 8, 1996, pp. 1233–1244.

http://dx.doi.org/10.1016/0031-3203(95)00160-3

[29] D. Patra and J. Mridula, "Featured Based Segmentation of Color Textured Images Using GLCM and Markov Ran-dom Field Model," *World Academy of Science, Engineering and Technology*, Vol. 53, No. 5, 2011, pp. 108-113.